\documentclass[runningheads]{llncs}
\usepackage{graphicx}
\usepackage{amsmath,amssymb} 
\usepackage{color}

\usepackage{floatrow}
\usepackage{amsmath}
\usepackage{amssymb}
\usepackage{bm}
\usepackage{url}
\usepackage[linesnumbered,ruled]{algorithm2e}
\usepackage{subcaption}
\usepackage{booktabs}
\usepackage{tabularx}
\usepackage{mathtools}

\usepackage{hyperref}
\usepackage{xcolor} 

\hypersetup{     
	colorlinks,     
	linkcolor={red!70!black},     
	citecolor={blue!95!black},
	urlcolor={black}
}
 
\newcommand*{\Scale}[2][4]{\scalebox{#1}{$#2$}}%

\begin{document}
\pagestyle{headings}
\mainmatter
\def\ECCVSubNumber{3240}

\title{Informative Sample Mining Network for \\ Multi-Domain Image-to-Image Translation}
\titlerunning{INIT for Multi-Domain Image-to-Image Translation}

\author{Jie Cao\inst{1,3}\orcidID{0000-0001-6368-4495} \and Huaibo Huang\inst{1,3}\orcidID{0000-0001-5866-2283} \and
Yi Li\inst{1,3}\orcidID{0000-0002-2856-7290} \and Ran He\inst{1,2,3\,*}\orcidID{0000-0002-3807-991X} \and Zhenan Sun\inst{1,2,3}\orcidID{0000-0003-4029-9935}}

\institute{
Center for Research on Intelligent Perception and Computing, NLPR, CASIA \and
Center for Excellence in Brain Science and Intelligence Technology, CAS \and
School of Artificial Intelligence, University of Chinese Academy of Sciences \\
\email{\{jie.cao, huaibo.huang, yi.li\}@cripac.ia.ac.cn}, \email{\{rhe, znsun\}@nlpr.ia.ac.cn}
}

\maketitle
\begin{abstract}
The performance of multi-domain image-to-image translation has been significantly improved by recent progress in deep generative models. Existing approaches can use a unified model to achieve translations between all the visual domains. However, their outcomes are far from satisfying when there are large domain variations. In this paper, we reveal that improving the sample selection strategy is an effective solution. To select informative samples, we dynamically estimate sample importance during the training of Generative Adversarial Networks, presenting Informative Sample Mining Network. We theoretically analyze the relationship between the sample importance and the prediction of the global optimal discriminator. Then a practical importance estimation function for general conditions is derived. Furthermore, we propose a novel multi-stage sample training scheme to reduce sample hardness while preserving sample informativeness. Extensive experiments on a wide range of specific image-to-image translation tasks are conducted, and the results demonstrate our superiority over current state-of-the-art methods.
\end{abstract}

\section{Introduction}
\label{sec:1}

\begin{figure}[ht]
	\centering
	\includegraphics[width=0.8\textwidth]
	{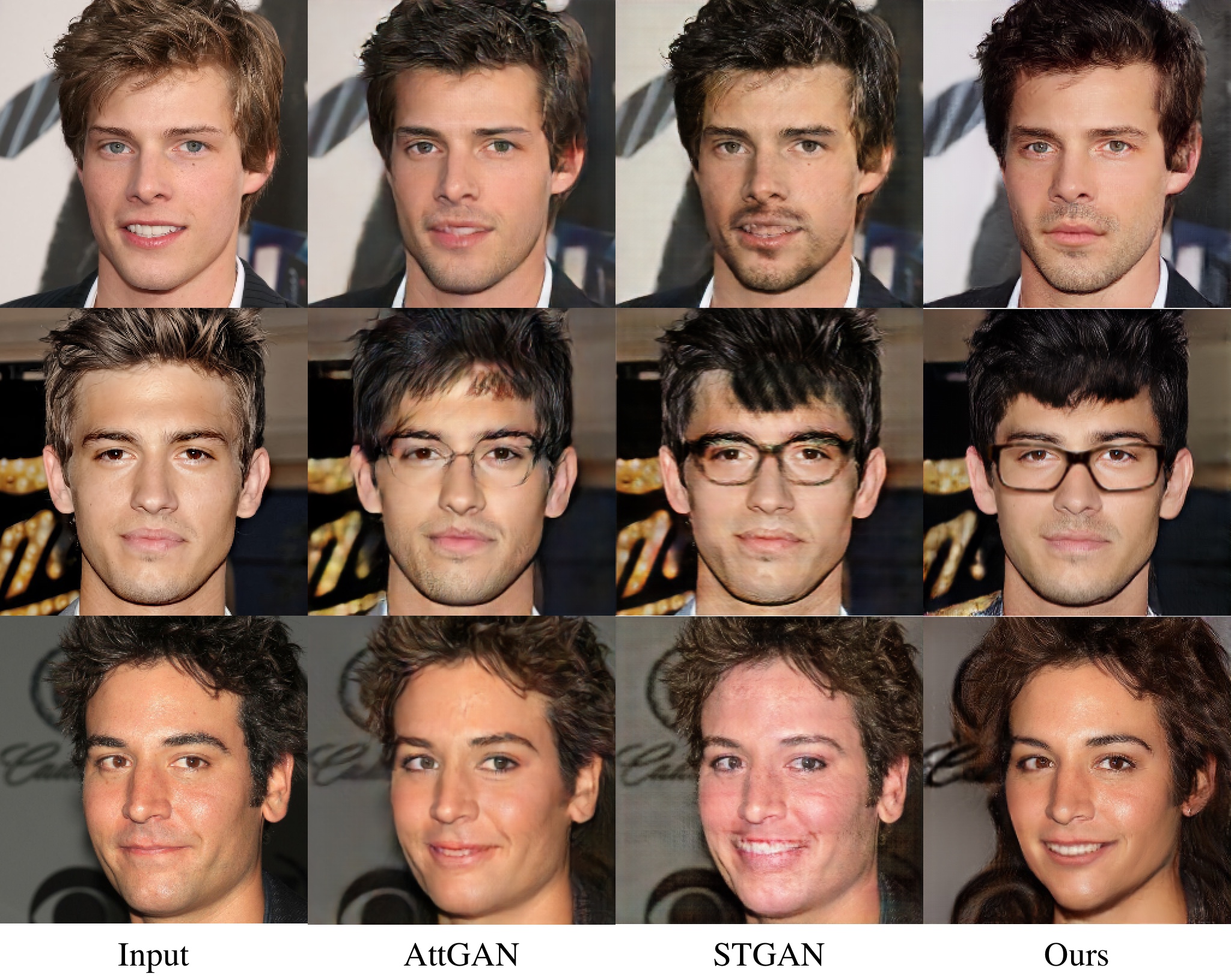}
	\caption{Facial attribute transfer results of different methods. From top to bottom, the target attributes of each row are ``\textit{beard}, \textit{no\_smiling}, \textit{black\_hair}'', ``\textit{bangs}, \textit{eyeglasses}, \textit{black\_hair}'', and  ``\textit{female}, \textit{smiling}, \textit{brown\_hair}'', respectively. While AttGAN \cite{he2019attgan} and STGAN \cite{liu2019stgan} show degraded results in these challenging cases, our approach achieves effective attribute transfer and maintains the realism of texture details.}
	\label{fig:1}
\end{figure}

Multi-domain image-to-image translation (I2I) aims at learning the mappings between visual domains. These domains can be instantiated by a set of attributes, each of which represents a meaningful visual property. Since each possible combination of the attributes specifies a unique domain \cite{patel2015visual}, the total domain number can be huge in practical applications. For instance, if there are 10 independent binary attributes, we need to handle the translations among 1024 visual domains, which is far more challenging than the two-domain translation. Fortunately, thanks to the recent advances in deep generative models \cite{goodfellow2014generative,kingma2013auto}, current methods \cite{choi2018stargan,he2019attgan,liu2019stgan,wu2019relgan} can achieve multi-domain I2I by a single model. However, these methods only produce promising results when translating images within similar visual domains, e.g., changing human hair color. The translations between domains with large semantic discrepancies are still not well addressed yet.

To further illustrate the limitation of existing approaches, we take the task of facial attribute transfer as a prime example. Fig.~\ref{fig:1} shows the results of some challenging translations. Due to the large gap between the source and the target domain, it is difficult to transfer target attributes without impairing visual realism. Even the current state-of-the-art methods \cite{he2019attgan,liu2019stgan} produce degraded results, although they can address most of the easy translations (we will show these cases in the following experiments). This phenomenon indicates that existing methods mainly focus on easy cases during training but neglect hard ones.

We argue that effective sample selection strategies greatly help to address this problem. Sample selection is within the scope of deep metric learning, which contributes to many computer vision tasks. The studies in deep metric learning \cite{huang2016local,schroff2015facenet,yuan2017hard,zheng2019hardness} point out that a large fraction of training samples may satisfy the loss constraints, providing no progress for model learning. That is, the vast majority of samples are too easy, so their contributions to our training are only marginal. Unfortunately, current multi-domain I2I methods merely adopt the naive random sample selection that treats all training samples equally and thus selects the easy ones mostly. It intuitively hinders training efficiency and consequently leads to the degradation discussed above.

In this paper, we propose \textbf{In}formative sample m\textbf{i}ning ne\textbf{t}work (INIT) to enhance training efficiency and improve performance in multi-domain I2I tasks. Concretely, we integrate Importance Sampling into the generation framework under Generative Adversarial Networks (GAN). Adversarial Importance Weighting is proposed to select informative samples and assign them greater weight. We derive the weighting function based on the assumption that the global optimal discriminator is known. Then we consider more general conditions and introduce the guidance from the prior model to rescale the importance weight. Furthermore, we propose Multi-hop Sample Training to avoid the potential problems \cite{schroff2015facenet,wu2017sampling,zheng2019hardness} in model training caused by sample mining. Based on the principle of divide-and-conquer, we produce target images by multiple hops, which means the image translation is decomposed into several separated steps. On the one hand, our training scheme preserves sample informativeness. On the other hand, step-by-step training ensures that the generator can learn complex translations. Combining with Adversarial Importance Weighting and Multi-hop Sample Training, our approach can probe and then fully utilize informative training samples.

To verify the effectiveness of our approach, we conduct experiments on facial attribute transfer, season transfer, and edge\&photo transfer. We make extensive comparisons with current state-of-the-art multi-domain I2I methods. The experimental results demonstrate our improvements in both attribute transfer and content preservation.

Our contributions can be summarized as follows:

\begin{itemize}
	
	\item[$\bullet$] We analyze the importance of sample selection in image-to-image translation and propose Informative Sample Mining Network.
	
	\item[$\bullet$] We propose Adversarial Importance Weighting, which integrates Importance Sampling into GAN, to achieve effective training sample mining.
	
	\item[$\bullet$] We propose Multi-hop Sample Training to reduce the hardness of the probed informative samples, making them easy to train.
	
	\item[$\bullet$] We provide extensive experimental results on facial attribute transfer, season transfer, and edge\&photo transfer, showing our superiority over existing approaches.
	
\end{itemize}

\section{Related work}
\label{sec:2}

\textbf{Image-to-Image Translation.} Recent advances in deep generative models \cite{goodfellow2014generative,kingma2013auto,oord2016pixel} have brought much progress in the field of image-to-image translation~\cite{choi2018stargan,deng2020reference,he2019attgan,isola2017image,lample2017fader,liu2019stgan,liu2017unsupervised,wu2019relgan,zhu2017unpaired}. At the early stage, the studies are focused on translations between two visual domains. Zhu~et al. have done pioneering works on learning the translations with paired data \cite{isola2017image} and unpaired data \cite{zhu2017unpaired}. FaderNet~\cite{lample2017fader} disentangles the salient information in the latent space to control attribute intensity. UNIT~\cite{liu2017unsupervised} combines Variational AutoEncoder~\cite{kingma2013auto} with GAN, and present high-quality results on unsupervised translation tasks. Later on, a lot of efforts are made to deal with the multi-domain condition. StarGAN~\cite{choi2018stargan} is the first unified model that produces visually plausible multi-domain translation results. AttGAN~\cite{he2019attgan} introduces an attribute-aware constraint as well as a reconstruction-based regularization to achieve ``only change what you want''. Following AttGAN, Liu~et al. \cite{liu2019stgan} propose a novel selective transfer unit to enhance image quality. RelGAN~\cite{wu2019relgan} introduces the notion of relative attribute and employs multiple discriminators to improve both attribute translation and interpolation. Different from previous approaches, we make improvements from a new perspective: we probe informative samples during training, making our network aware of the most challenging cases.

\bigbreak

\noindent\textbf{Deep Metric Learning.} Deep metric learning aims at learning good representations. The core idea is to narrow the distances of similar images in the embedding space and enlarge the distances of dissimilar ones. Existing works mainly focus on how to choose proper loss functions and sample selection strategies. In the studies of loss function, contrastive loss~\cite{hu2014discriminative} and triplet loss~\cite{wang2014learning} are the most representative works. The two losses are widely adopted and extended by successive methods. Huang~et al.~\cite{huang2016local} explore the structure of quadruplets. Wang~et al.~\cite{wang2017deep} improve triplet loss by introducing a third-order geometry relationship. Sample selection strategies have also been widely studied. For example, hard negative sample mining~\cite{simo2015discriminative} is proposed to replace the random sample selection in the contrastive loss. FaceNet~\cite{schroff2015facenet} first adopts semi-hard negative mining within a batch for face recognition. Harwood~et al.~\cite{harwood2017smart} utilize approximate nearest neighbor search to select harder samples adaptively. Recently proposed methods \cite{duan2018deep,zhao2018adversarial} introduce adversarial learning to generate potentially informative samples to train the model. At present, sample selection strategies have been adopted in many computer vision tasks, including image classification \cite{law2013quadruplet}, face recognition \cite{cao2020domain,guo2020learning,hu2014discriminative,schroff2015facenet}, and person re-identification \cite{yu2018hard}. In this work, we show that sample strategy is also important to image generation but rarely studied. To this end, we propose a novel sampling strategy specifically for multi-domain I2I.

\bigbreak

\noindent\textbf{Importance Sampling.} Importance Sampling (IS) \cite{evans1995methods} is a method for estimating properties of a target distribution $p_{data}(\bm{x})$ which is difficult to sample from directly. Samples are instead drawn from a proposal distribution $p_g(\bm{x})$ that over-weights the important region. IS is essential for many statistical theories, including Bayesian inference \cite{evans1995methods} and sequential Monte Carlo methods \cite{oh1993integration,veach1995optimally}. Applying IS, we can draw samples from $p_g(\bm{x})$ to estimate $\mathbb{E}_{p_{data}}\left[ \mathcal{L}(\bm{X}) \right]$ for any known function $\mathcal{L}$. Specifically, we have 
\begin{equation}
\label{eq:1}
\mathbb{E}_{p_{data}} \left[ \mathcal{L}(\bm{X}) \right] = \int \frac{ p_{data}(\bm{x})}{p_g(\bm{x})} \mathcal{L} (\bm{x}) p_g(\bm{x}) \mathrm{d} \bm{x} = \mathbb{E}_{p_g} \left[ \frac{p_{data}(\bm{X})}{p_g(\bm{X})} \mathcal{L}(\bm{X}) \right],
\end{equation}
where for any $\bm{x}$ in the sample space, we have $p_g(\bm{x})>0$ whenever $p_g(\bm{x}) \cdot p_{data}(\bm{x})\neq 0$. The likelihood ratio $\frac{p_{data}(\bm{x})}{p_g(\bm{x})}$ is also referred to as the importance weight. In the following section, we will integrate IS into the GAN-based multi-domain I2I methods.

\section{Proposed Method}
\label{sec:3}

We consider visual domains characterized by an n-dimensional binary attribute vector $\bm{a}=\left[ a_1,  a_2, \cdots, a_n \right]^T$, where each bit $a_i$ represents a meaningful visual attribute. We build Informative Sample Mining Network to learn all the mappings between these domains from unpaired training data. Our network takes a source image~$\bm{x}^s$ with a target attribute~$\bm{a}^t$ to produce the corresponding fake target image~$\bm{x}^t$.

\begin{figure}[t]
	\centering
	\includegraphics[width=\textwidth]
	{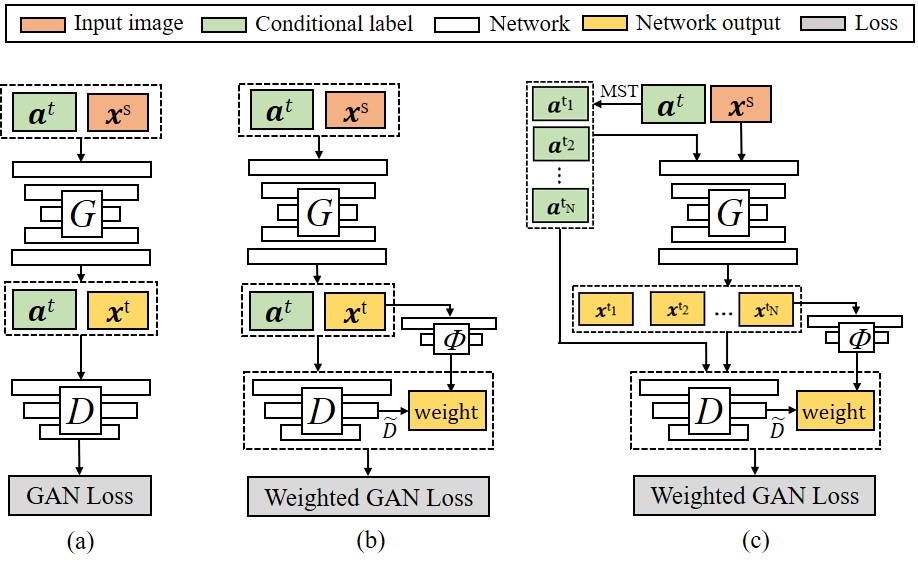}
	\caption{A illustration about our framework. (a) is our backbone model, which is a Conditional GAN. (b) is the improved version with the proposed AIW (see Section~\ref{sec:31}). Our full model, which combines AIW and MST (see Section~\ref{sec:32}), is represented as (c).} 
	\label{fig:211}
\end{figure}

We adopt a Conditional GAN \cite{mirza2014conditional} as our backbone model, which is illustrated in Fig.~\ref{fig:211}~(a). That is, we train the generator and make the generator distribution $p_g$ to capture the true data distribution $p_{data}$. Meanwhile, a discriminator tries to distinguish the real data from the synthesized fake data. The generator and the discriminator are trained jointly by optimizing the adversarial loss, which can be written as: 
\begin{equation}
\label{eq:2} 
\min _{G} \max _{D} \mathcal{L}=\overbrace{\mathbb{E}_{\bm{x}^s,\bm{a}^s \sim p_{data}} \left[\log D(\bm{x}^s, \bm{a}^s)\right]}^{\mathcal{L}_{data}} + \overbrace{\mathbb{E}_{\bm{x}^t,\bm{a}^t \sim p_{g}} [\log (1 - D(\bm{x}^t, \bm{a}^t))]}^{\mathcal{L}_{g}},
\end{equation}
where $\bm{x}^t=G(\bm{x}^s, \bm{a}^t)$. For brevity, We use $G$ and $D$ to denote the generator and the discriminator, respectively. The inputs of $G$ and $D$ are the concatenation of an image and an attribute vector. We apply spatial replication on the attribute vector, making the sizes of the image and the attribute vector matched.

\subsection{Adversarial Importance Weighting}
\label{sec:31}

In this section, we describe how to improve the sampling strategy of our backbone model, proposing Adversarial Importance Weighting. We will first consider the situation where we have the global optimal discriminator and then discuss the generalized situation.

To emphasize the contributions of informative samples, we improve the estimation of $\mathcal{L}_{g}$ by introducing an importance weight for each fake sample $\bm{x}^t$. Specifically, we aim to calculate the weight $\frac{p_{data}(\bm{x}^t)}{p_{g}(\bm{x}^t)}$, which is introduced in Eq.~\ref{eq:1}. To this end, we need to find a solution to make the weight computable. Recall the proposition made by Goodfellow~et~al.~\cite{goodfellow2014generative}: for any fixed $G$ and any sample point $\bm{x}^t$, we have
\begin{equation}
\label{eq:3} 
D^{*}(\bm{x}^{t})=\frac{p_{data}(\bm{x}^{t})}{p_{data}(\bm{x}^{t})+p_{g}(\bm{x}^{t})},
\end{equation}
where $D^{*}$ is the global optimal discriminator. To reveal the relation between the discriminator and the importance weight, let $D^{*}(\bm{x}^{t})=S(\widetilde{D}^{*}(\bm{x}^{t}))$, where $S$ denotes the sigmoid function. That is, we have
\begin{equation} 
\label{eq:31} 
D^{*}(\bm{x}^t)=\frac{1}{1+e^{-\widetilde{D}^{*}(\bm{x}^t)}}.
\end{equation}

Combining Eq.~\ref{eq:3} and Eq.~\ref{eq:31}, we can derive that $\frac{p_{data}(\bm{x}^t)}{p_{g}(\bm{x}^t)} = e^{\widetilde{D}^{*}(\bm{x}^t)}$. Hence, the weighted $\mathcal{L}_{g}$ can be formulated as:
\begin{equation} 
\label{eq:4} 
\mathcal{L}_{g} = \mathbb{E}_{\bm{x}^t,\bm{a}^t \sim p_{g}} [e^{\widetilde{D}^{*}(\bm{x}^t)} \cdot \log (1 - D(\bm{x}^t, \bm{a}^t))].
\end{equation}

Eq.~\ref{eq:4} indicates that a greater $\widetilde{D}^{*}(\bm{x}^t)$ brings $\bm{x}^t$ a bigger sample weight, which means $\bm{x}^t$ is more informative. In the meantime, a greater $ \widetilde{D}^{*}(\bm{x}^t)$ also indicates $\bm{x}^t$ is harder to distinguish for the discriminator. Hence, similar to existing sample selection strategies \cite{schroff2015facenet,wu2017sampling,zheng2019hardness}, \textbf{mining the informative samples in GAN means finding the hard fake samples}.

Now we consider the practical situation, where we cannot get the optimal discriminator. A straightforward way to sidestep the need of $D^{*}$ is replacing it with $D$. However, $D$ may not provide accurate estimation if it is too far away from the optimality. Hence, we aim at measuring how close $D$ is to $D^{*}$. Inspired by the fact \cite{goodfellow2014generative} that $D$ is close to $D^{*}$ when $p_g$ is similar to $p_{data}$, we propose a heuristic metric. Concretely, we first project each training batch $\{\bm{x}^s_1,~\bm{x}^s_2,\cdots,\bm{x}^s_n\}$ and the corresponding generated results $\{\bm{x}^t_1,~\bm{x}^t_2,\cdots,\bm{x}^t_n\}$ onto a hypersphere whose radius is $r$ by a pre-trained embedding model $\phi$. Then, we can calculate the distance matrix $\bm{E}$, where $e_{ij}=||\phi(\bm{x}^s_i) - \phi(\bm{x}^t_j)||$, i.e., the Euclidean distance between $\bm{x}^s_i$ and $\bm{x}^t_j$ in the embedding space. We define that
\begin{equation}
\label{eq:42}
\Scale[0.8]{\Delta}l = \overbrace{(\sum\limits_{i,j} e_{ij} - \sum\limits_{i} e_{i})}^{\Scale[0.8]{\Delta}l_n} - \overbrace{\sum\limits_{i} e_{i}}^{\Scale[0.8]{\Delta}l_p} = \sum\limits_{i,j} e_{ij} - 2\cdot\text{trace}(\bm{E}),
\end{equation}
where $\Scale[0.8]{\Delta}l_p$ denotes the sum of the distances between the relative image pairs (e.g., $e_{11}$,  $e_{33}$), and $\Scale[0.8]{\Delta}l_n$ denotes the sum of the distances between the permuted image pairs (e.g., $e_{12}$,  $e_{31}$). Since only the source and the generated images that form a relative pair have the same content information, $\Scale[0.8]{\Delta}l_p$ should be as small as possible, and $\Scale[0.8]{\Delta}l_n$ should be as large as possible. In the optimal situation, we have $\Scale[0.8]{\Delta}l^*_n = 2r$ and $\Scale[0.8]{\Delta}l^*_p = 0$. Hence, $\Scale[0.8]{\Delta}l^{*}$ is a determined constant namely $2r$. In practical conditions, we can calculate $(\Scale[0.8]{\Delta} l^{*} - \Scale[0.8]{\Delta} l)$ to measure how close our network is to the global optimality. Formally, introducing the scale factor $(\Scale[0.8]{\Delta} l^{*} - \Scale[0.8]{\Delta} l)$ into the importance weight, we propose Adversarial Importance Weighting, which can be formulated as:
\begin{equation}
\label{eq:5}
\text{AIW}(\bm{x}^t) = \Vert 1 + \left(\Scale[0.8]{\Delta} l^{*} - \Scale[0.8]{\Delta} l\right) \Vert^2 \cdot e^{\widetilde{D}(\bm{x}^t)}.
\end{equation}

We have introduced AIW into our backbone model, as depicted in Fig.~\ref{fig:211}~(b). Accordingly, the formula of $\mathcal{L}_{g}$ is updated to:
\begin{equation} 
\label{eq:53} 
\mathcal{L}_{g} = \mathbb{E}_{\bm{x}^t,\bm{a}^t \sim p_{g}} [\text{AIW}(\bm{x}^t) \cdot \log (1 - D(\bm{x}^t, \bm{a}^t))].
\end{equation}

\subsection{Multi-hop Sample Training}
\label{sec:32}

\begin{figure}[t]
	\centering
	\includegraphics[width=0.8\textwidth]{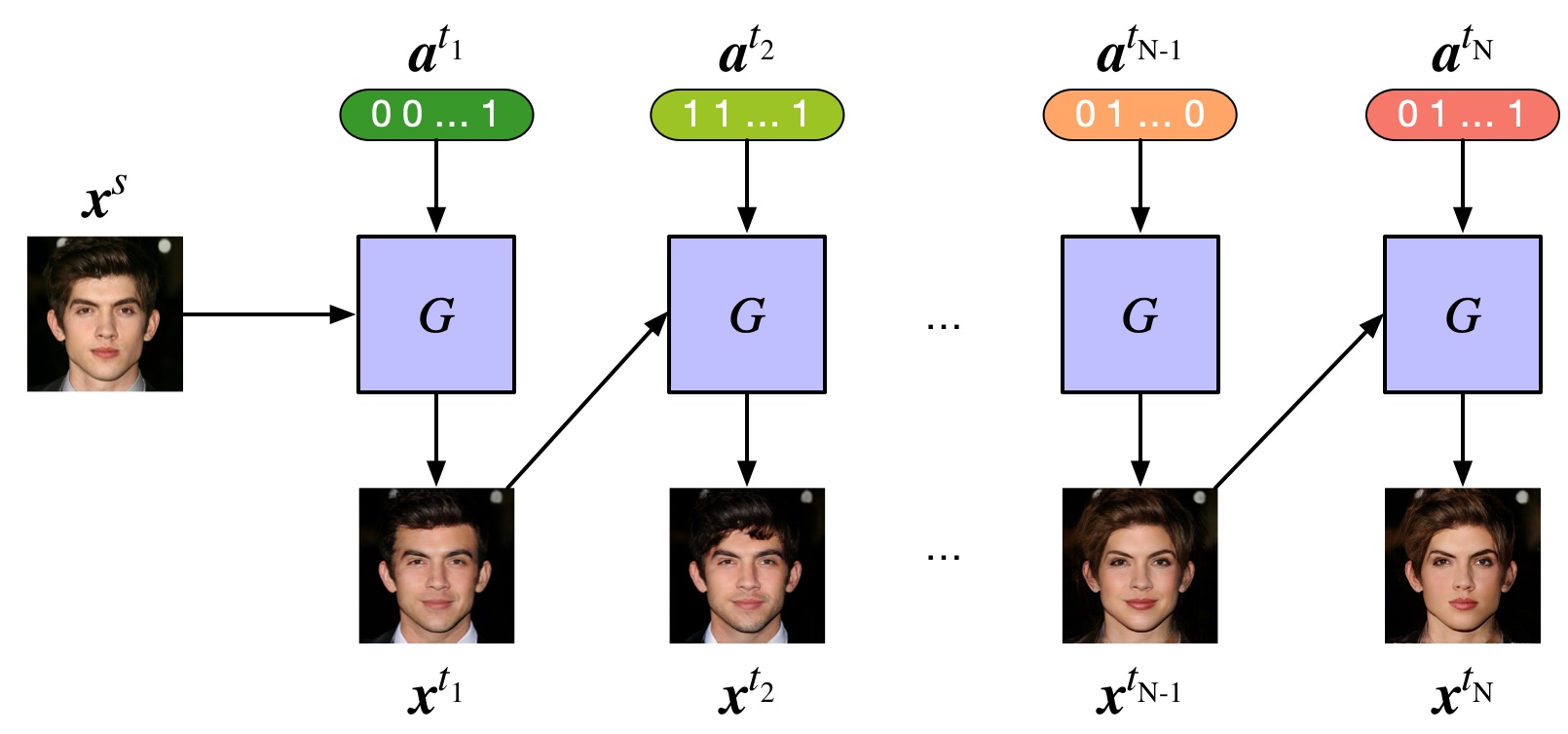}
	\caption{An illustration about $N$-hop target image generation. After $N$ times of translations, the generator produces the $N$-hop target image $\bm{x}^{t_N}$.}	
	\label{fig:23}
\end{figure}

The proposed AIW makes the discriminator aware of hard samples and thus strengthens its power. However, since $D$ and $G$ are rivals during training, the training of $G$ may become problematic due to the superior $D$. To address this issue, we introduce Multi-hop Sample Training which reduces sample hardness for the generator in a divide-and-conquer manner.


Let ``hop'' denote translation time a model takes to produce the target result. Fig.~\ref{fig:23} provides a visual illustration, and here we give a formal definition:
\begin{equation}
\label{eq:6}
\bm{x}^{t_N} = \overbrace{G(~\cdots~G(G}^{\text{N}}(\bm{x}^s,\bm{a}^{t_1}),~\bm{a}^{t_2}),~\cdots),~\bm{a}^{t_N}),
\end{equation}
where generator transforms $\bm{x}^s$ into $\bm{x}^{t_N}$ via $N$ separate steps, and $\bm{x}^{t_N}$ is denoted as $N$-hop target image. We define $\{\bm{a}^{t_1},\bm{a}^{t_2},~\cdots,~\bm{a}^{t_{N-1}}\}$ as intermediate attributes, and $\{\bm{x}^{t_1},\bm{x}^{t_2},~\cdots,~\bm{x}^{t_{N-1}}\}$ are inferred to as intermediate images.

Previous approaches only consider the situation where $N=1$. Consequently, some complex transformations may be too hard to learn for the generator. However, any complex transformation can be shrunk step-by-step (e.g., transfer multiple target attributes one-by-one), and it suffices to construct a feasible solution step-by-step. Therefore, we propose to reduce sample hardness by generating the target image in a multi-hop manner. Concretely, we introduce Multi-hop Sample Training, which considers \{1-hop, 2-hop, $\cdots$, $N$-hop\} target images. During training, we calculate losses on both the target images and the intermediate images, providing supervision information for each single step in the multi-hop image generation. Equipping the backbone model with MST, we update the formula of $\mathcal{L}_{g}$ to:
\begin{equation} 
\label{eq:64} 
\mathcal{L}_{g} = \sum\limits_{n=1}^{N} \mathbb{E}_{p_{g}} [\text{AIW}(\bm{x}^t) \cdot \mathbb{E}_{p_{\text{n-hop}}} [\sum\limits_{i=1}^{n} \log (1 - D(\bm{x}^{t_i}, \bm{a}^{t_i}))]],
\end{equation}
where $\{\bm{a}^{t_1},\bm{a}^{t_2},~\cdots,~\bm{a}^{t_{n-1}}\} \sim p_{\text{n-hop}}$ ($n=1,2,\cdots,N$), and we draw these intermediate attributes randomly in our experiments. Note that we also add AIW, which is proposed in Eq.~\ref{eq:5}, to this equation.

\subsection{Implementation Details}
\label{sec:33}

\begin{algorithm}[!t]
	
	Pretrain the embedding model $\phi$
	
	Initialize the generator $G$ and the discriminator $D$ 
	
	\For{the number of training epochs}
	{
		Darw a training sample batch
		
		$G$ forward propagates, producing 1-hop target images
		
		$D$ forward propagates
		
		Calculate sample importance weights by Eq.~\ref{eq:5}
		
		Draw intermediate attributes 
		
		$G$ forward propagates, producing multi-hop target images
		
		$D$ forward propagates
		
		Calculate $\mathcal{L}_{data} = \mathbb{E}_{\bm{x}^s,\bm{a}^s \sim p_{data}} \left[\log D(\bm{x}^s, \bm{a}^s)\right]$ 
		
		Calculate $\mathcal{L}_g$ by Eq.~\ref{eq:64}
		
		Calculate $\mathcal{L} = \mathcal{L}_{data} + \mathcal{L}_g$
		
		Optimize $G$ by minimizing $\mathcal{L}$
		
		Optimize $D$ by maximizing $\mathcal{L}$
	}
	\caption{Training algorithm of INIT}
	\label{alg:1}
\end{algorithm}

Combining AIW and MST, our full model is able to select informative samples and then train them effectively. The complete diagram is depicted in Fig.~\ref{fig:211}~(c). During training, we follow the methodology of classical GAN \cite{goodfellow2014generative} and optimize $G$ and $D$ iteratively. We first produce 1-hop target images, just like the previous approaches. Then, we estimate the importance weight by Eq.~\ref{eq:5}. Next, we draw intermediate attributes and produce multi-hop target images. Finally, we update model parameters by optimizing the weighted adversarial loss on multi-hop images. The training process is summarized in Algorithm~\ref{alg:1}. 

In our experiments, we adopt 2-hop MST. We build a fully convolutional network as our generator and use a patch discriminator similar to \cite{zhu2017unpaired}. The pre-trained VGG \cite{simonyan2014very} is employed as the embedding model. Specifically, we use VGGFace \cite{parkhi2015deep} for facial attribute transfer. We optimize model parameters by Adam optimizer \cite{kingma2014adam} with $\beta_1 = 0.5$, $\beta_2 = 0.999$, and a learning rate of 1e-4. During testing, our generator directly produces the 1-hop target images as the output, which is the same as existing I2I methods.

\section{Experiments}
\label{sec:4}

To validate the effectiveness of our approach, we perform extensive experiments on facial attribute transfer, season transfer, and edge\&photo transfer. We produce $256 \times 256$ results and train our model as well as other competing models for 100 epochs. Our batch size is set to 16. In the following part, we first describe the datasets in our experiments (Section~\ref{sec:41}). Then we make comparisons with existing methods and report experimental results (Section~\ref{sec:42},~\ref{sec:43},~and~\ref{sec:44}). Finally, we present an ablation study (Section~\ref{sec:45}).

\begin{figure}[t]
	\centering
	\includegraphics[width=\textwidth]
	{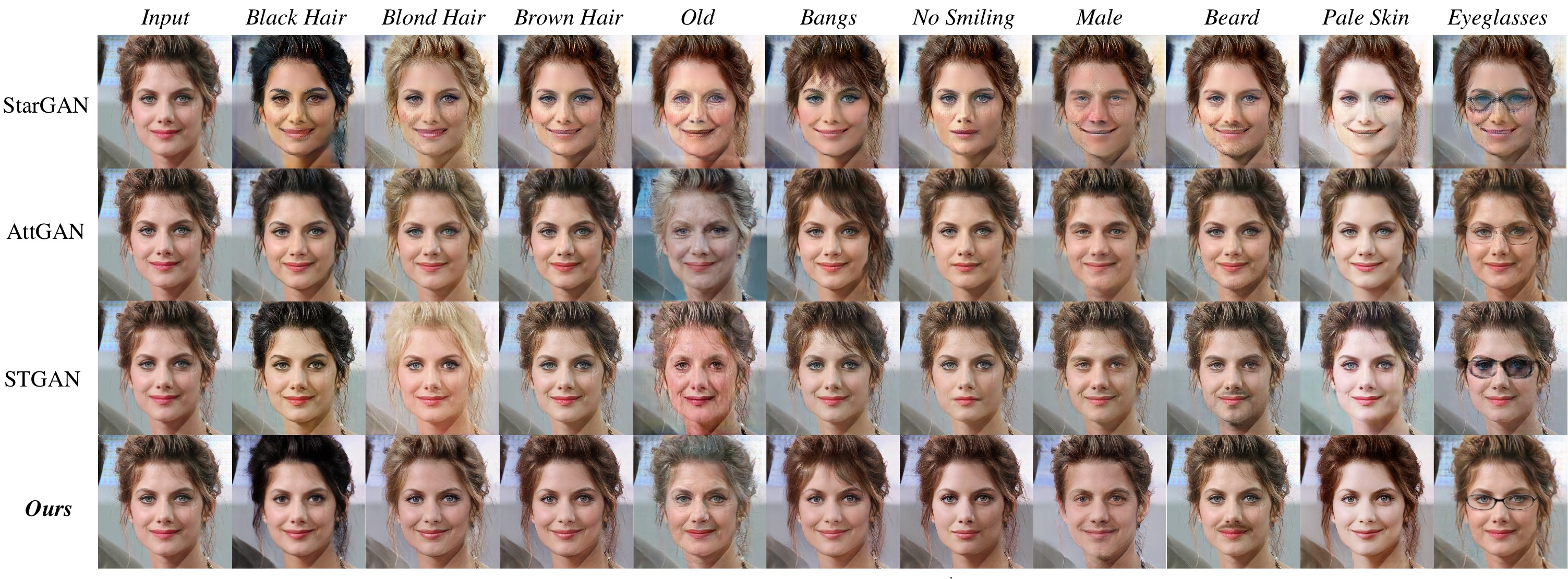}
	\caption{Visual examples of single facial attribute editing results. From top to bottom, the rows are results of StarGAN~\cite{choi2018stargan}, AttGAN~\cite{he2019attgan}, STGAN~\cite{liu2019stgan}, and our INIT.}
	\label{fig:3}
\end{figure}

\begin{table}[t]
	\centering
	\caption{The comparisons of the classification accuracy (\%) \cite{wu2019relgan} of each attribute (higher is better) and Fr\'echet Inception Distance \cite{heusel2017gans} (FID, lower is better) on facial attribute transfer.}
	\label{tab:1}
	\scalebox{1.1}
	{
		\begin{tabular}{cccccc}
			\toprule
			& StarGAN \cite{choi2018stargan} & AttGAN \cite{he2019attgan} & STGAN \cite{liu2019stgan} & Ours  & Real Data \\
			\midrule
			Hair Color & 91.02 &         93.10  & 92.45 & \textbf{94.47} & 96.12 \\
			Aging      & 92.38 &         95.41  & 95.22 & \textbf{97.90} & 98.42 \\
			Bangs      & 87.97 &         91.03  & 91.84 & \textbf{93.26} & 93.67 \\
			Smile      & 85.53 &         90.47  & 87.41 & \textbf{90.94} & 91.00 \\
			Gender     & 90.72 & \textbf{96.77} & 94.76 &         96.57  & 98.25 \\
			Beard      & 87.90 &         93.53  & 93.09 & \textbf{95.58} & 95.34 \\
			Skin Color & 89.35 &         92.65  & 94.03 & \textbf{94.69} & 94.22 \\
			Eyeglasses & 93.38 &         96.44  & 96.09 & \textbf{98.46} & 99.31 \\
			\midrule
			FID        & 19.28 &         13.62  & 15.94 & \textbf{11.16} &   -   \\
			\bottomrule
		\end{tabular}
	}
\end{table}

\begin{figure}[t]
	\centering
	\includegraphics[width=\textwidth, height=330pt]
	{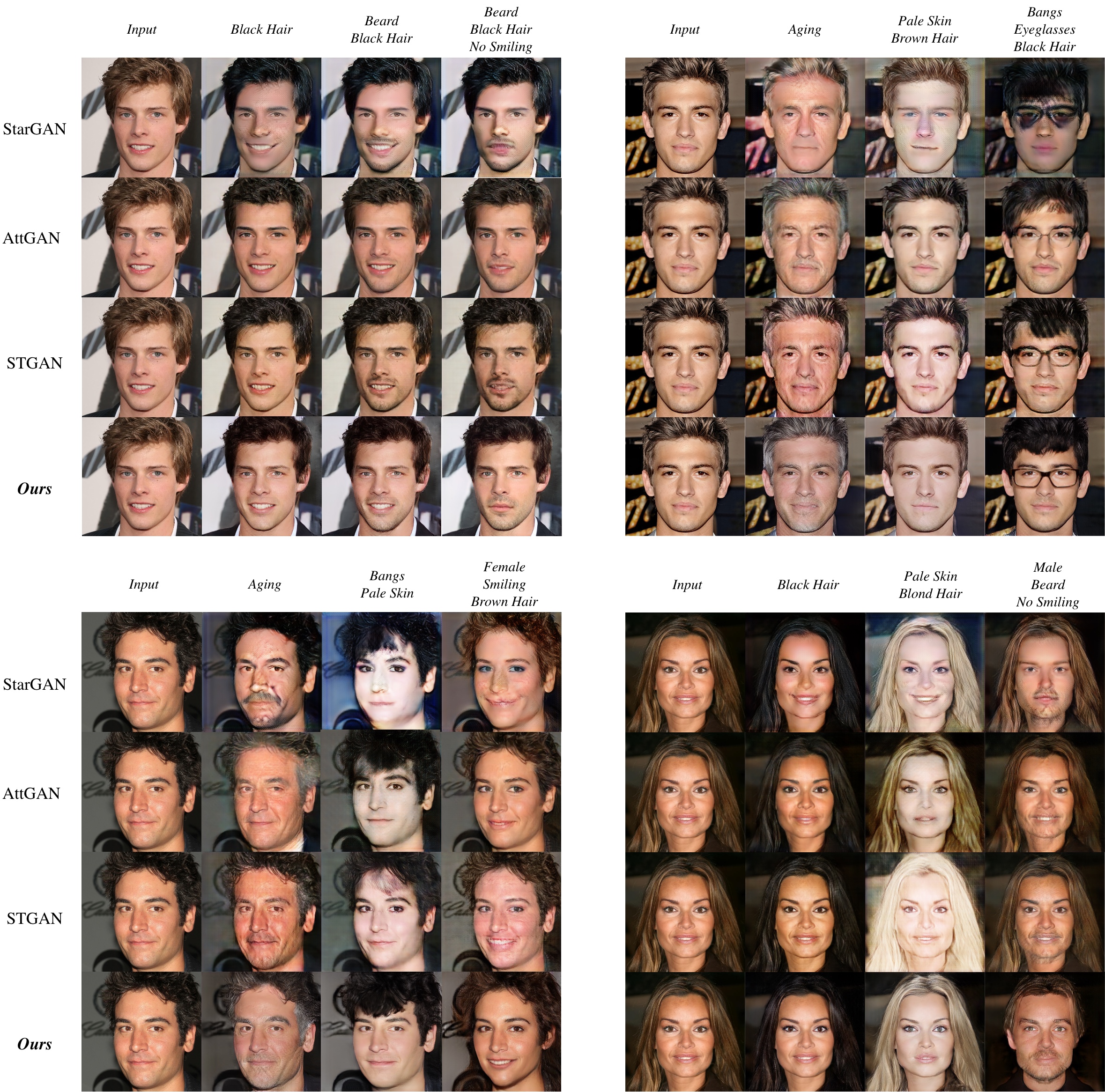}
	\caption{Visual examples of multiple facial attribute editing, which is a very challenging case in multi-domain I2I. Please zoom in for better visualization. We make comparisons with StarGAN~\cite{choi2018stargan}, AttGAN~\cite{he2019attgan}, STGAN~\cite{liu2019stgan}.}
	\label{fig:4}
\end{figure}

\subsection{Dataset}
\label{sec:41}

\textbf{CelebA}~\cite{liu2015deep} is the largest publicly available dataset for multi-domain I2I tasks at present. There are annotations of 40 binary attributes for each image. In our experiment, we use the high-quality version, CelebA-HQ~\cite{karras2017progressive}, for facial attribute transfer. We choose the following 10 attributes to construct the attribute vector: \textit{Black\_Hair}, \textit{Blond\_Hair}, \textit{Brown\_Hair}, \textit{Bangs}, \textit{Smiling}, \textit{Male}, \textit{No\_Beard}, \textit{Pale\_Skin}, and \textit{Eyeglasses}. We randomly select 300 images as the testing set and use all the remaining images for training.

\textbf{Yosemite Flickr Dataset}~\cite{zhu2017unpaired} consists of 1,200 winter photos and 1,540 summer photos of Yosemite National Park. It is widely used for season transfer, which is an unpaired I2I problem. In our experiment, we follow the training and testing data divisions of CycleGAN~\cite{zhu2017unpaired}.

\textbf{Edge2Photo Dataset}~\cite{eitz2012humans} contains photos of 250 categories of objects and the corresponding edges. Pix2pix~\cite{isola2017image} first uses the shoes category for edge-to-photo transfer. We follow the training and testing data divisions in Pix2pix to train our model.

\subsection{Facial Attribute Transfer}
\label{sec:42}

We compare with StarGAN~\cite{choi2018stargan}, AttGAN~\cite{he2019attgan}, and STGAN~\cite{liu2019stgan} on facial attribute transfer. We reproduce their results by the released source codes. The training and testing data for all the methods are the same.

\begin{figure}[t]
	\centering
	\begin{subfigure}[t]{0.48\textwidth}
		\includegraphics[width=\textwidth]{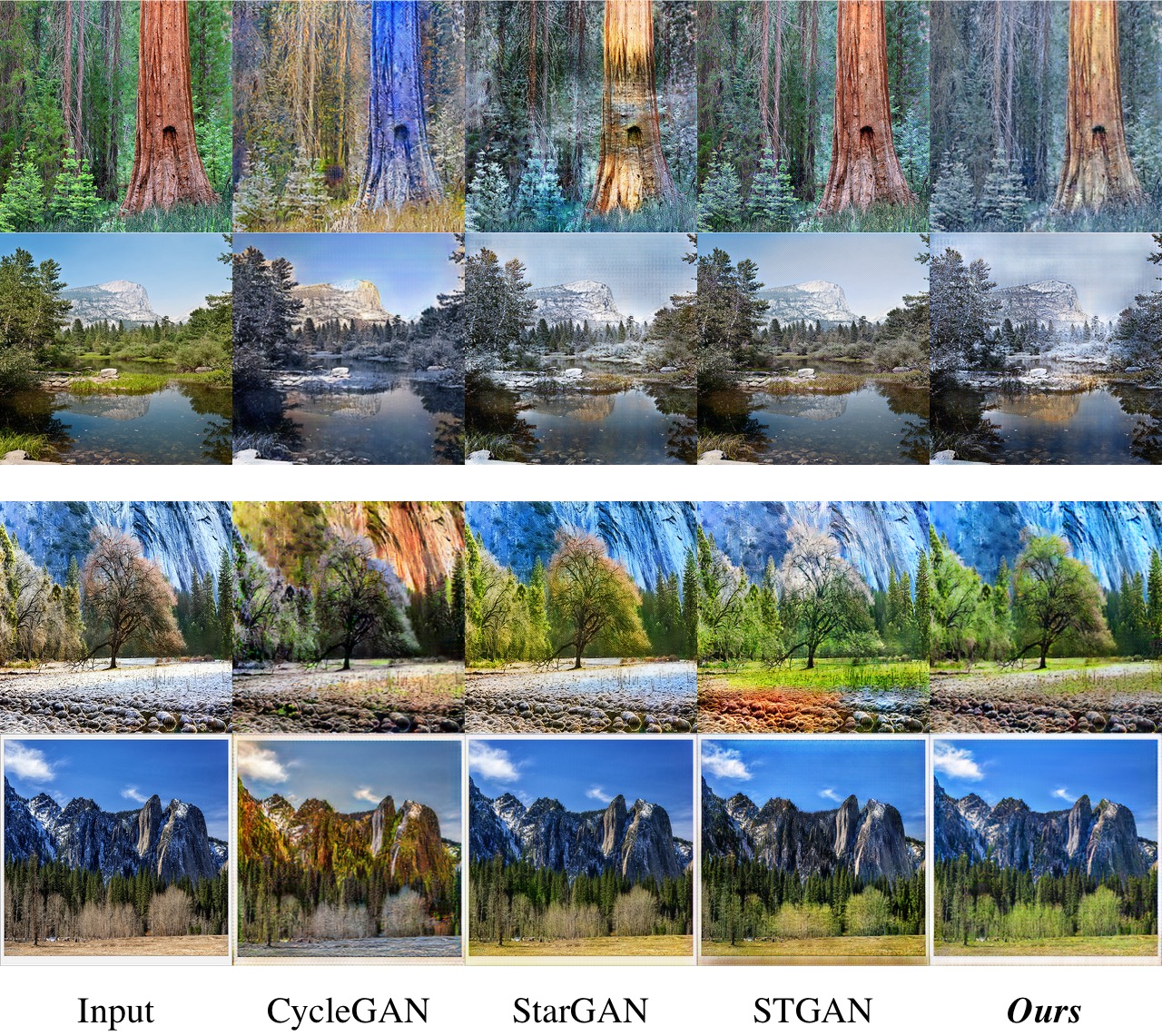}
		\caption{The two rows on the top are \textit{summer}$\rightarrow$\textit{winter}, and the two rows on the bottom are \textit{winter}$\rightarrow$\textit{summer}. We make comparisons with CycleGAN \cite{zhu2017unpaired}, StarGAN \cite{choi2018stargan}, and STGAN \cite{liu2019stgan}.}
		\label{fig:51}
	\end{subfigure}
	~
	\begin{subfigure}[t]{0.48\textwidth}
		\includegraphics[width=\textwidth]{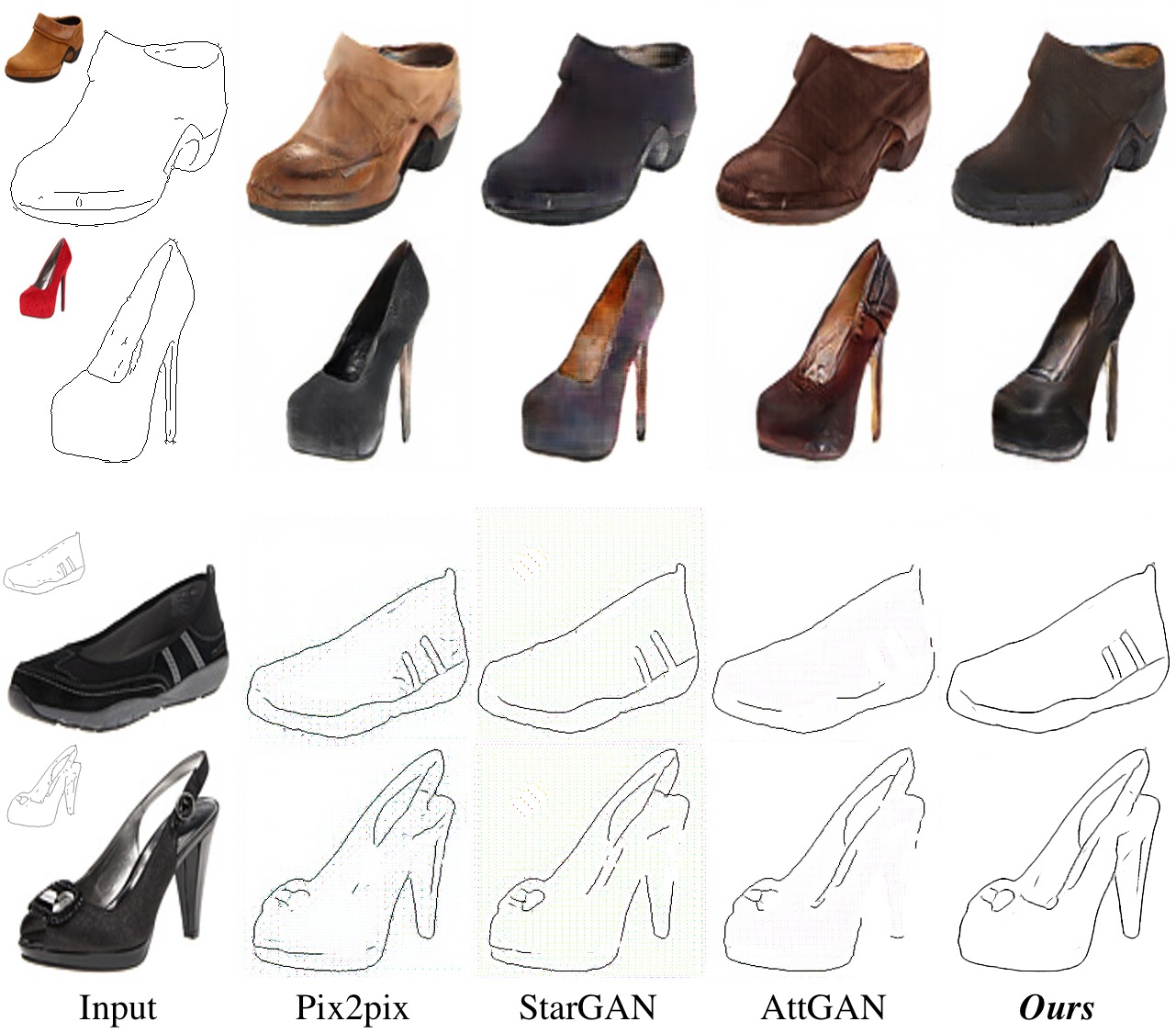}
		\caption{The top two rows are \textit{edge}$\rightarrow$\textit{shoes}, and the bottom two rows are \textit{shoes}$\rightarrow$\textit{edge}. We compare with Pix2pix \cite{isola2017image}, StarGAN \cite{choi2018stargan}, and AttGAN \cite{he2019attgan}. The ground truths are on the upper left corner of the inputs.}
		\label{fig:52}
	\end{subfigure}
\caption{Visual examples of (a) season transfer and (b) edge\&photo transfer.}
\label{fig:5}
\end{figure}

\begin{table}[t]
	\centering
	\caption{The comparisons of FID and the percent of user votes (higher is better). For (a) season transfer, the FID is reported as ``summer/winter''. For (b) edge\&photo transfer, the FID is reported as ``photo/edge''.}
	\scalebox{1.1}{
		\begin{tabular}{ccccc}
			\toprule
			\multicolumn{5}{c}{(a) \textbf{Season Transfer}} \\
			\midrule
			Metric &~CycleGAN~\cite{zhu2017unpaired}~&~StarGAN~\cite{choi2018stargan}~&~STGAN~\cite{liu2019stgan}~& Ours \\
			\midrule
			FID    & 48.40/49.39 & 44.84/52.16 & 45.61/50.37 & \textbf{40.09}/\textbf{44.91} \\
			Vote Percent & 3.73\%        & 11.07\%        & 21.52\%        & \textbf{63.68\% } \\
			\midrule
			\multicolumn{5}{c}{(b) \textbf{Edge\&Photo Transfer}} \\
			\midrule
			Metric &~Pix2pix~\cite{isola2017image}~&~StarGAN~\cite{choi2018stargan}~&~AttGAN~\cite{he2019attgan}~& Ours \\
			\midrule
			FID    & 39.59/17.13 & 33.23/13.92 & 29.01/13.86 & \textbf{28.44}/\textbf{12.30}  \\
			Vote Percent  & 30.26\%        & 3.77\%         & 6.44\%         & \textbf{59.53\%} \\
			\bottomrule
		\end{tabular}
	}
	\label{tab:2}
\end{table}

Fig.~\ref{fig:3} shows single attribute transfer results. Given an input image, each method produces 10 transformed images. For each output result, one specific attribute is toggled. For the relatively easy tasks like changing hair color, all the methods produce plausible results. By contrast, when dealing with the challenging ones like aging, our method produces more realistic results. In general, our INIT can achieve effective attribute transfer and outperform other methods.

We argue that \textbf{multiple attribute transfer} should also be emphasized. Transferring multiple attributes is at least no easier than transferring a single attribute, and the number of possible combinations is significantly larger. Hence, the hard cases are mainly from the multiple attribute conditions. We report comparison results of multiple facial attribute transfer in Fig.~\ref{fig:1} and Fig.~\ref{fig:4}. In these cases, keeping visual quality and achieving effective attribute transfer become far more challenging. However, our INIT can still produce the desired results. Thanks to the weighting strategies, our method pays more attention to the hard cases and therefore yields better performance. Our superiority provides strong evidence on the effectiveness of sample selection.

We also calculate Fr\'echet Inception Distance (FID) \cite{heusel2017gans} and the classification accuracy \cite{wu2019relgan} to make objective comparisons. Lower FID is better since it means that the Wasserstein distance between the real distribution and the generated distribution is smaller. The classification accuracy reflects the effectiveness of attribute transfer, and thus higher is better. Following \cite{liu2019stgan} , we train a Resnet-18 \cite{he2016deep} as the classifier and calculate the accuracy on the transformed results. To train this classifier, we use the same data division of CelebA-HQ~\cite{karras2017progressive} as the division for our generation tasks. In Table~\ref{tab:1}, we summarize FID and the classification accuracy of each class. It can be observed that our method has the best performance, indicating our improvements in visual realism and attribute transfer.

\subsection{Season Transfer}
\label{sec:43}

For season transfer, we make comparisons with CycleGAN~\cite{zhu2017unpaired}, StarGAN~\cite{choi2018stargan}, and STGAN~\cite{liu2019stgan}. Note that only CycleGAN trains a pair of networks to achieve \textit{summer}$\rightarrow$\textit{winter} and \textit{winter}$\rightarrow$\textit{summer}, respectively. The other approaches can achieve season transfer by a single model.

We summarize the visual examples of translation results in Fig.~\ref{fig:51}. We also calculate FID as an objective metric. Since real summer and winter photos have an apparent perceptual discrepancy, we calculate FID on the two seasons separately. Furthermore, we conduct a user study. We invite volunteers to select the best result among the transformed images from the four methods. All the testing images are compared, and we report the percent of votes for each method. The quantitative comparison results are summarized in the upper part of Table~\ref{tab:2}. The comparison on FID indicates that our approach favorably outperforms the competing methods, and we obtain the majority of the user votes.

\subsection{Edge\&Photo Transfer}
\label{sec:44}

In this subsection, our method is compared with Pix2pix~\cite{isola2017image}, StarGAN~\cite{choi2018stargan}, and AttGAN~\cite{he2019attgan} on edge\&photo transfer. Pix2pix needs to learn edge-to-photo and photo-to-edge separately, and the other methods can deal with the two translations simultaneously. Since we have paired data in this dataset, we add the L1 distance loss~\cite{isola2017image} in the pixel space, which is useful for paired I2I tasks. Note that we also add the L1 loss for the other competing methods to make fair comparisons.

We report the examples of translation results and the ground truth in Fig.~\ref{fig:52}. Similar to season transfer, we calculate FID and conduct user study, the results of which are reported in the lower part of Table~\ref{tab:2}. Learning edge-to-photo and photo-to-edge as two separate tasks brings Pix2pix obvious advantages, but our method still has the best performance. Compared with StarGAN and AttGAN, our method produces more plausible results, showing stronger generalization ability for paired I2I tasks.

\subsection{Ablation Study}
\label{sec:45}

\begin{table}[t]
	\centering
	\caption{Comparison results of different variations of our method. $w/$ and $w/o$ are the abbreviations of ``with'' and ``without'', respectively. Our full model is equivalent to the variation (c).}
	\scalebox{1.1}{
	\begin{tabular}{cllccc}
		\toprule
		Model~&~AIW~&~MST~&~Hop Number~&~FID~&~Mean Acc \\
		\midrule
		(a) & $w/o$ & $w/$ & 2 & 18.91 & 92.86 \\
		(b) & $w/$  & $w/$ & 3 & 11.23 & 94.98 \\
		(c) & $w/$  & $w/$ & 2 & \textbf{11.16} & \textbf{95.08} \\
		(d) & $w/$  & $w/o$ & 1 & 14.52 & 93.78 \\
		(e) & $w/o$ & $w/o$ & 1 & 21.38 & 91.44 \\
		\bottomrule
	\end{tabular}
	}
	\label{tab:3}
\end{table}

In this subsection, we conduct an ablation study to verify the effectiveness of AIW and MST. To this end, we implement several variations of our approach and evaluate them on facial attribute transfer. Concretely, we consider the following variations: (a) INIT without any importance sampling schemes, (b-d) INIT with $n$-hop sample training, where $n=3,2,1$, respectively. (e) INIT that removes both AIW and MST, i.e., simply a conditional GAN \cite{mirza2014conditional}. Note that our full model is equivalent to variation (c).

We use the same experiment setting and train these variations for the same number of iterations. Note that variations with a smaller hop number will have more iterations for training new samples since they have fewer intermediate results to optimize. Tabel~\ref{tab:3} shows comparison results on quantitative metrics, and please refer to our Supplementary for visual examples. Through the ablation study, we can verify the following two points:

\textbf{Mining informative samples plays an important role.} Without the important sampling scheme, the performances of variations (a) and (e) drop sharply. Even when we double the training iterations of variation (e), its performance is still obviously inferior. Hence, merely taking more training time is not an effective option.

\textbf{The optimal choice is 2-hop sample training.} As the hop number increases, more intermediate samples are drawn during training. It means that we pay more attention to reduce the sample hardness for the generator. However, as indicated by Eq.~\ref{eq:64}, it also means 1-hop samples contribute less to the loss function. Note that during testing, the evaluations are based on 1-hop target images. Hence, a larger hop number does not guarantee better performance in practice.

\section{Conclusion}

In this paper, we propose to integrate Importance Sampling into a GAN-based model, resulting in Adversarial Importance Weighting for high-quality multi-domain image-to-image translation. Furthermore, Multi-hop Sample Training subtly reduces sample hardness while preserving sample informativeness. Thanks to the improvements in training efficiency, our approach achieves effective translation even when dealing with a large number of challenging visual domains. We conduct extensive experiments on practical tasks, including facial attribute transfer, season transfer, and edge\&photo transfer. The results consistently demonstrate our superiority over existing methods.

\mbox{}\\

\noindent\textbf{Acknowledgement.} This work is funded by the National Natural Science Foundation of China (Grant No. U1836217), Beijing Natural Science Foundation (Grant No. JQ18017) and Youth Innovation Promotion Association CAS (Grant No. Y201929).

\clearpage
%
%


\begin{thebibliography}{10}
\providecommand{\url}[1]{\texttt{#1}}
\providecommand{\urlprefix}{URL }
\providecommand{\doi}[1]{https://doi.org/#1}

\bibitem{cao2020domain}
Cao, D., Zhu, X., Huang, X., Guo, J., Lei, Z.: Domain balancing: Face
  recognition on long-tailed domains. In: CVPR (2020)

\bibitem{choi2018stargan}
Choi, Y., Choi, M., Kim, M., Ha, J.W., Kim, S., Choo, J.: {StarGAN}: Unified
  generative adversarial networks for multi-domain image-to-image translation.
  In: CVPR (2018)

\bibitem{deng2020reference}
Deng, Q., Cao, J., Liu, Y., Chai, Z., Li, Q., Sun, Z.: Reference guided face
  component editing (2020)

\bibitem{duan2018deep}
Duan, Y., Zheng, W., Lin, X., Lu, J., Zhou, J.: Deep adversarial metric
  learning. In: CVPR (2018)

\bibitem{eitz2012humans}
Eitz, M., Hays, J., Alexa, M.: How do humans sketch objects? In: SIGGRAPH
  (2012)

\bibitem{evans1995methods}
Evans, M., Swartz, T., et~al.: Methods for approximating integrals in
  statistics with special emphasis on bayesian integration problems. Stat Sci
  (1995)

\bibitem{goodfellow2014generative}
Goodfellow, I., Pouget-Abadie, J., Mirza, M., Xu, B., Warde-Farley, D., Ozair,
  S., Courville, A., Bengio, Y.: Generative adversarial nets. In: NeurIPS
  (2014)

\bibitem{guo2020learning}
Guo, J., Zhu, X., Zhao, C., Cao, D., Lei, Z., Li, S.Z.: Learning meta face
  recognition in unseen domains. In: CVPR (2020)

\bibitem{harwood2017smart}
Harwood, B., Kumar, B., Carneiro, G., Reid, I., Drummond, T., et~al.: Smart
  mining for deep metric learning. In: ICCV (2017)

\bibitem{he2016deep}
He, K., Zhang, X., Ren, S., Sun, J.: Deep residual learning for image
  recognition. In: CVPR (2016)

\bibitem{he2019attgan}
He, Z., Zuo, W., Kan, M., Shan, S., Chen, X.: {AttGAN}: Facial attribute
  editing by only changing what you want. TIP  (2019)

\bibitem{heusel2017gans}
Heusel, M., Ramsauer, H., Unterthiner, T., Nessler, B., Klambauer, G.,
  Hochreiter, S.: {GANs} trained by a two time-scale update rule converge to a
  {Nash} equilibrium. In: NeurIPS (2017)

\bibitem{hu2014discriminative}
Hu, J., Lu, J., Tan, Y.P.: Discriminative deep metric learning for face
  verification in the wild. In: CVPR (2014)

\bibitem{huang2016local}
Huang, C., Loy, C.C., Tang, X.: Local similarity-aware deep feature embedding.
  In: NeurIPS (2016)

\bibitem{isola2017image}
Isola, P., Zhu, J.Y., Zhou, T., Efros, A.A.: Image-to-image translation with
  conditional adversarial networks. In: CVPR (2017)

\bibitem{karras2017progressive}
Karras, T., Aila, T., Laine, S., Lehtinen, J.: Progressive growing of {GANs}
  for improved quality, stability, and variation. In: ICLR (2018)

\bibitem{kingma2014adam}
Kingma, D.P., Ba, J.: Adam: A method for stochastic optimization. In: ICLR
  (2015)

\bibitem{kingma2013auto}
Kingma, D.P., Welling, M.: Auto-encoding variational bayes. In: ICLR (2014)

\bibitem{lample2017fader}
Lample, G., Zeghidour, N., Usunier, N., Bordes, A., Denoyer, L., Ranzato, M.:
  Fader networks: Manipulating images by sliding attributes. In: NeurIPS (2017)

\bibitem{law2013quadruplet}
Law, M.T., Thome, N., Cord, M.: Quadruplet-wise image similarity learning. In:
  ICCV (2013)

\bibitem{liu2019stgan}
Liu, M., Ding, Y., Xia, M., Liu, X., Ding, E., Zuo, W., Wen, S.: {STGAN}: A
  unified selective transfer network for arbitrary image attribute editing. In:
  CVPR (2019)

\bibitem{liu2017unsupervised}
Liu, M.Y., Breuel, T., Kautz, J.: Unsupervised image-to-image translation
  networks. In: NeurIPS (2017)

\bibitem{liu2015deep}
Liu, Z., Luo, P., Wang, X., Tang, X.: Deep learning face attributes in the
  wild. In: ICCV (2015)

\bibitem{mirza2014conditional}
Mirza, M., Osindero, S.: Conditional generative adversarial nets. arXiv
  preprint arXiv:1411.1784  (2014)

\bibitem{oh1993integration}
Oh, M.S., Berger, J.O.: Integration of multimodal functions by monte carlo
  importance sampling. J AM STAT ASSOC  (1993)

\bibitem{oord2016pixel}
Oord, A.v.d., Kalchbrenner, N., Kavukcuoglu, K.: Pixel recurrent neural
  networks. In: ICML (2016)

\bibitem{parkhi2015deep}
Parkhi, O.M., Vedaldi, A., Zisserman, A., et~al.: Deep face recognition. In:
  BMVC (2015)

\bibitem{patel2015visual}
Patel, V.M., Gopalan, R., Li, R., Chellappa, R.: Visual domain adaptation: A
  survey of recent advances. Signal Process. Mag.  (2015)

\bibitem{schroff2015facenet}
Schroff, F., Kalenichenko, D., Philbin, J.: Facenet: A unified embedding for
  face recognition and clustering. In: CVPR (2015)

\bibitem{simo2015discriminative}
Simo-Serra, E., Trulls, E., Ferraz, L., Kokkinos, I., Fua, P., Moreno-Noguer,
  F.: Discriminative learning of deep convolutional feature point descriptors.
  In: ICCV (2015)

\bibitem{simonyan2014very}
Simonyan, K., Zisserman, A.: Very deep convolutional networks for large-scale
  image recognition. arXiv preprint arXiv:1409.1556  (2014)

\bibitem{veach1995optimally}
Veach, E., Guibas, L.J.: Optimally combining sampling techniques for monte
  carlo rendering. In: SIGGRAPH (1995)

\bibitem{wang2017deep}
Wang, J., Zhou, F., Wen, S., Liu, X., Lin, Y.: Deep metric learning with
  angular loss. In: ICCV (2017)

\bibitem{wang2014learning}
Wang, J., Song, Y., Leung, T., Rosenberg, C., Wang, J., Philbin, J., Chen, B.,
  Wu, Y.: Learning fine-grained image similarity with deep ranking. In: CVPR
  (2014)

\bibitem{wu2017sampling}
Wu, C.Y., Manmatha, R., Smola, A.J., Krahenbuhl, P.: Sampling matters in deep
  embedding learning. In: ICCV (2017)

\bibitem{wu2019relgan}
Wu, P.W., Lin, Y.J., Chang, C.H., Chang, E.Y., Liao, S.W.: {RelGAN}:
  Multi-domain image-to-image translation via relative attributes. In: ICCV
  (2019)

\bibitem{yu2018hard}
Yu, R., Dou, Z., Bai, S., Zhang, Z., Xu, Y., Bai, X.: Hard-aware point-to-set
  deep metric for person re-identification. In: ECCV (2018)

\bibitem{yuan2017hard}
Yuan, Y., Yang, K., Zhang, C.: Hard-aware deeply cascaded embedding. In: ICCV
  (2017)

\bibitem{zhao2018adversarial}
Zhao, Y., Jin, Z., Qi, G.j., Lu, H., Hua, X.s.: An adversarial approach to hard
  triplet generation. In: ECCV (2018)

\bibitem{zheng2019hardness}
Zheng, W., Chen, Z., Lu, J., Zhou, J.: Hardness-aware deep metric learning. In:
  CVPR (2019)

\bibitem{zhu2017unpaired}
Zhu, J.Y., Park, T., Isola, P., Efros, A.A.: Unpaired image-to-image
  translation using cycle-consistent adversarial networks. In: ICCV (2017)

\end{thebibliography}
\end{document}